\documentclass[sigconf]{acmart}
\AtBeginDocument{%
  }

\setcopyright{none}
\copyrightyear{}
\acmDOI{}
\acmConference[]{}{}{}
\acmISBN{}

\usepackage{doi}

\begin{document}

\title{ZhuLong: Execution-Grounded LLM Agent for EDA Scripting with Offline API Self-Exploration}

\author{%
  \begin{tabular}{c}
    Yang Liu \quad Shiwei Hou \quad Xiyuan Chen \quad Yu Wang \quad Sen Yuan \\[1pt]
    {\small \{paul.liu, shiwei.hou\}@cxmt.com \quad chenxiyuan@mail.ustc.edu.cn} \\[4pt]
    Qirui Gan \quad Shao You \quad Feifan Chen \quad Wencheng Li \quad Shuyang Hu \\[1pt]
    Yongzhou Liu \quad Emma Xia \quad Xiaojing Lu \quad Hao Wang \quad Fan Xu \quad Yanfeng Li
  \end{tabular}
}

\affiliation{%
  \institution{Changxin Memory Technologies, Inc.}
  \country{China}
}

\renewcommand{\shortauthors}{Liu et al.}

\begin{abstract}
    EDA scripting with tool-specific, often undocumented APIs remains a long-tail 
    bottleneck that existing LLMs fail to address. This paper presents 
    \textbf{ZhuLong}, an execution-grounded LLM coding agent for PyAether and 
    SKILL that combines API retrieval, documentation inspection, and sandbox 
    execution via unified MCP tools, augmented by an offline \textbf{API 
    self-exploration mechanism} that infers undocumented API behaviors through 
    counterfactual experimentation.
    
    We evaluate ZhuLong on \textbf{EDA-Eval-PyAether}, a benchmark of 158
    real-world tasks with assertion-based execution, where the complete system
    achieves \textbf{78.5\%} Pass@1 in the commercial Empyrean Aether
    environment, substantially outperforming a pure LLM baseline (\textbf{23.6\%}). Ablation studies identify sandbox execution as the dominant
    performance driver (41.2 pp drop when removed), with the self-exploration
    mechanism contributing an additional 3.2 pp accuracy gain and a 22.1\%
    reduction in per-task tool calls. On 20 interactive tasks involving unsaved
    layouts and schematics, ZhuLong achieves 60.0\% Pass@1 for PyAether and
    50.0\% for SKILL. 
\end{abstract}



\maketitle

\section{Introduction}
\label{sec:introduction}

Large language models (LLMs) have demonstrated strong capabilities in general-purpose code generation \cite{vaswani2017attention, openai2023gpt4, yang2025qwen3}. However, applying LLMs to EDA scripting—where engineers routinely manipulate in-memory design databases, create layout instances, route schematics, and query netlists—remains challenging. First, EDA APIs are long-tailed and tool-specific; many rarely appear in public training corpora. Second, correctness cannot be verified statically: it depends on observable side effects in a real EDA environment, such as modified layouts or updated schematics. Third, without execution feedback, static generation cannot diagnose errors or iteratively refine incorrect code \cite{chen2024dawning, blocklove2024eda}. Existing LLM-based EDA work has mainly focused on HDL generation \cite{thakur2023verigen, liu2023rtlcoder}, while scripting with real tool APIs—especially with incomplete documentation and sandbox execution—remains underexplored.

This paper presents \textbf{ZhuLong}, an execution-grounded LLM coding agent for PyAether and SKILL. Named after a mythical creature whose opened eyes bring light, ZhuLong aims to illuminate opaque EDA APIs and automate tool-specific scripting workflows. Built on Cline-CLI \cite{cline2024cline}, ZhuLong extends three MCP tools \cite{lee2025autoeda}—\texttt{search\_apis}, \texttt{get\_api\_details}, and \texttt{run\_code}—that form a closed loop: retrieve candidate APIs, inspect their documentation, execute code in the EDA environment with execution feedback, and iteratively refine based on observed errors or outputs. All tools accept a unified \texttt{lang} parameter, supporting both PyAether and SKILL with identical agent logic. To further address incomplete or outdated documentation, ZhuLong introduces an offline \textbf{API self-exploration mechanism} that proactively explores undocumented API behaviors in the sandbox via \textbf{counterfactual experimentation}—deliberately testing alternative parameter values, observing execution outcomes, inferring constraints and usage patterns, and storing the discovered information as enhanced API documentation for future queries.

The main contributions are:
\begin{itemize}
    \item \textbf{First execution-grounded agent for commercial EDA scripting.} We present the first system that validates execution feedback in PyAether and SKILL via sandbox execution, delivering a +43 pp accuracy gain over static RAG.
    \item \textbf{Offline API self-exploration for EDA.} A complementary mechanism that contributes +3.2 pp accuracy gain and reduces the average number of tool calls per trace by 22.1\% through counterfactual experimentation on undocumented APIs. To our knowledge, this is the first mechanism that proactively discovers and codifies undocumented API behaviors in the EDA domain.
    \item \textbf{EDA-Eval-PyAether: the first public benchmark for PyAether scripting.} A benchmark of 158 real-world tasks with assertion-based execution, enabling rigorous evaluation of LLM-generated EDA scripts. The complete ZhuLong system achieves 78.5\% Pass@1 on this benchmark.
\end{itemize}
\section{Background and Motivation}
\label{sec:background}

We target two widely-used EDA scripting interfaces: PyAether for Empyrean Aether \cite{empyrean_pyaether} and SKILL for Cadence Virtuoso \cite{cadence2024virtuoso}. Despite their prevalence, scripting for both environments poses challenges beyond what generic LLM code generators can address, and the community lacks basic infrastructure to measure progress.

\textbf{EDA scripting is fundamentally different.} Prior work on LLM code generation has succeeded in domains where correctness is verified by unit tests \cite{chen2021evaluating, austin2021program}. EDA scripting presents three distinct challenges: (i) APIs are long-tailed, tool-specific, and poorly documented; (ii) correctness depends on side effects in a live, stateful database, not return values; and (iii) no benchmark exists to evaluate whether generated scripts actually work.

\textbf{Prior work stops short of execution.} Recent efforts like LayoutCopilot \cite{liu2025layoutcopilot}, ChaTCL \cite{rui2025chatcl}, and RAG-EDA \cite{pu2024rag} combine LLMs with static knowledge bases. Yet they all share a critical limitation: they do not execute generated code. Correctness is assessed by human judgment \cite{liu2025layoutcopilot, rui2025chatcl} or retrieval quality \cite{pu2024rag}, not by whether the script actually modifies a layout. Consequently, errors go undiagnosed, and the gap between "looks plausible" and "actually works" remains unbridged.

\textbf{The root problem: no benchmark, no progress.} Without a standardized benchmark, the community cannot systematically compare approaches or measure progress—each prior work uses ad-hoc tasks with different protocols. Recent concurrent benchmarks for Tcl/Innovus flows \cite{xu2026iscript, li2026pdagent} confirm the growing recognition of this gap, yet target different toolchains. A dedicated PyAether/SKILL benchmark remains absent.

\textbf{Our approach.} ZhuLong addresses the technical challenges through sandbox execution feedback—closing the loop from generation to execution to refinement—complemented by an offline \textbf{API self-exploration mechanism}. Crucially, we introduce \textbf{EDA-Eval-PyAether}, the first standardized benchmark for PyAether scripting. Together, these contributions establish both the methodology and the measurement infrastructure for advancing LLM-based EDA scripting.
\section{System Design}
\label{sec:system}

\subsection{Overall Architecture}
\label{sec:overall-architecture}

\begin{figure*}[t]
\centering
\includegraphics[width=0.9\textwidth]{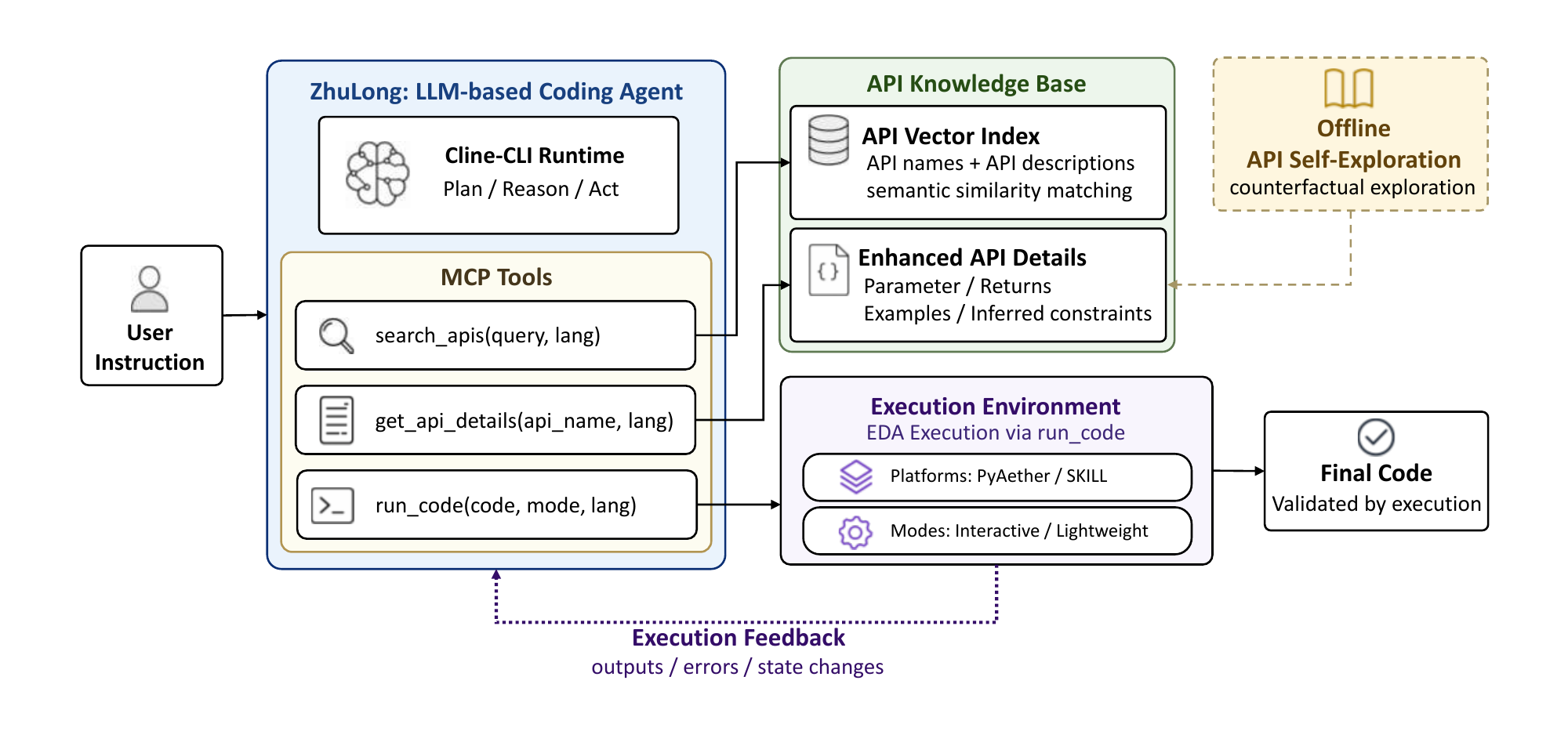}
\caption{ZhuLong system architecture. The agent interacts with API retrieval, enhanced documentation, and EDA execution via three unified MCP tools. Offline self-exploration augments the knowledge base before runtime, and execution feedback supports iterative refinement.}
\label{fig:architecture}
\end{figure*}

ZhuLong is built on Cline-CLI \cite{cline2024cline}, an open-source framework for autonomous coding agents, extended with EDA-specific retrieval, documentation, and execution capabilities. We choose Cline-CLI for its mature permission controls, human-in-the-loop support, and enterprise readiness. As shown in Figure~\ref{fig:architecture}, the system comprises three components: the LLM-based agent runtime, the API knowledge base, and the EDA execution environment. The agent interacts with these components through three MCP tools, while an offline \textbf{API self-exploration mechanism} pre-augments the API knowledge base before runtime.

\subsection{Three MCP Tools}
\label{sec:mcp-tools}

ZhuLong extends three MCP tools \cite{mcp2024protocol}, each accepting a unified \texttt{lang} parameter (\texttt{pyaether} or \texttt{skill}) to support both platforms with identical agent logic:

\begin{itemize}
    \item \texttt{search\_apis(query, lang)} retrieves relevant APIs via semantic and keyword-based search. We index APIs by name and description using BGE-M3 embeddings \cite{bge2024} with FAISS \cite{douze2024faiss} for efficient similarity search, returning the top-$K$ results ($K=5$ by default) with confidence scores and descriptions.
    
    \item \texttt{get\_api\_details(api\_name, lang)} fetches comprehensive documentation for a specified API, including parameter types, return values, and examples. The returned documentation is augmented by the offline API self-exploration mechanism (Section~\ref{sec:selfdoc}) when available, enriching incomplete specifications with constraints and usage patterns discovered via sandbox exploration.
    
    \item \texttt{run\_code(code, mode, lang)} executes the generated code in the EDA environment. It supports two modes: (1) \textbf{lightweight}—runs in a sandboxed environment without a GUI session (standalone PythonE for PyAether, or SKILL in batch mode for Virtuoso), capturing outputs, errors, and state changes for iterative refinement; (2) \textbf{interactive}—runs inside an open EDA session via the agent-EDA bridge (socket communication), enabling operations on unsaved layouts and schematics.
\end{itemize}

\subsection{Agent Workflow}
\label{sec:agent-workflow}

The agent iteratively refines its output through execution feedback \cite{cline2024cline}. A typical trajectory begins with reasoning over the user instruction, followed by API retrieval via \texttt{search\_apis} when needed. Candidate APIs are then inspected via \texttt{get\_api\_details}, and code is generated accordingly. The agent executes the code via \texttt{run\_code}, observes the result, and re-plans—revising the code, retrieving alternative APIs, or terminating. This loop continues until success or the agent determines further attempts are unlikely to succeed. The agent is not constrained to a fixed sequence; it may skip retrieval when it already knows the required API, or conduct multiple rounds of retrieval when initial results lack confidence.
\section{Offline API Self-Exploration Mechanism}
\label{sec:selfdoc}

API documentation in EDA tools is often incomplete, omitting parameter constraints, return structures, and error conditions. Prior work addresses documentation gaps through retrieval-based augmentation \cite{pu2024rag} or static knowledge graphs \cite{liu2025layoutcopilot}—both of which rely on what is already documented. When documentation is absent, these approaches provide no remedy.


ZhuLong takes a different approach: instead of waiting for documentation to exist, it \textbf{proactively discovers} API behaviors through offline counterfactual experimentation in the sandbox. For each target API, the self-exploration agent reads whatever documentation is available, identifies gaps—e.g., unspecified parameter ranges, undocumented return fields, or ambiguous error semantics—and generates targeted test scripts to probe them. It executes these scripts via \texttt{run\_code}, observes outcomes, and iteratively refines its hypotheses about what the API accepts and returns. The exploration agent autonomously decides what to test next and when to stop, based on its current confidence.


Concretely, the exploration agent constructs counterfactual hypotheses—e.g., whether a parameter accepts an empty list, whether a numeric argument allows negative values, or whether a return value contains a specific field—and designs minimal test cases to validate each. Execution outcomes (success, error, exception, or unexpected return structure) reveal actual constraints. Over multiple rounds, the agent builds a behavioral model of each API, capturing parameter constraints, return structures, error conditions, and side effects.


This offline process runs once per API; the enriched documentation is stored in the API knowledge base and served at runtime via \texttt{get\_api\_details}, incurring no runtime overhead. The resulting documentation is more complete than what is available in official sources, and directly reflects the API's actual behavior rather than its intended or assumed behavior.

\section{Benchmark Construction}
\label{sec:benchmark}

\subsection{Dataset: EDA-Eval-PyAether}

To evaluate LLM-based code generation for PyAether, we constructed \textbf{EDA-Eval-PyAether}, a benchmark of 158 tasks. Tasks were collected from four sources: API references (61.4\%), documentation examples (27.2\%), internal training materials (7.6\%), and anonymized CAD practical cases (3.8\%). Each task underwent manual review by EDA domain experts.

Table~\ref{tab:scenario-distribution} shows the distribution of tasks across different EDA scenarios. Layout and Schematic operations constitute the majority (70.2\%), reflecting the core focus of PyAether usage in physical design workflows.

\begin{table}[htbp]
\centering
\caption{Task distribution by scenario}
\label{tab:scenario-distribution}
\begin{tabular}{lcc}
\toprule
\textbf{Scenario} & \textbf{Tasks} & \textbf{Percentage} \\
\midrule
Layout & 64 & 40.5\% \\
Schematic & 47 & 29.7\% \\
Design Management & 24 & 15.2\% \\
Others & 23 & 14.6\% \\
\midrule
Total & 158 & 100\% \\
\bottomrule
\end{tabular}
\end{table}

\subsection{Evaluation Protocol}

Each task consists of a natural language prompt, a function signature, and assertion-based test code. A task is considered passed if the generated code executes in the PyAether sandbox without errors and all assertions pass. The primary metric is Pass@1 \cite{chen2021evaluating}:

\[
\text{Pass@1} = \frac{\text{\# tasks with all assertions passed}}{\text{Total tasks}} \times 100\%
\]
\section{Experiments}
\label{sec:experiments}

\subsection{Experimental Setup}
\label{sec:exp-setup}

\subsubsection{Research Questions}

This evaluation aims to answer four research questions: (RQ1) whether sandbox execution improves code generation accuracy over static generation; (RQ2) whether the self-exploration mechanism provides measurable benefits beyond sandbox execution alone, in terms of both accuracy and efficiency; (RQ3) whether vector-based retrieval outperforms grep-based keyword search for API discovery; and (RQ4) how different index construction strategies affect retrieval effectiveness.

\subsubsection{Benchmark and Evaluation Protocol}

We evaluate ZhuLong on \textbf{EDA-Eval-PyAether}, a benchmark of 158 real-world PyAether scripting tasks. Following the evaluation protocol defined in Section~\ref{sec:benchmark}, we adopt \textbf{Pass@1} (execution success rate) as the primary metric. During evaluation, the agent is allowed up to \textbf{2 re-planning rounds} per task (i.e., up to three complete runs), following the self-correction loop described in Section~\ref{sec:agent-workflow}; timeout is set to 2500 seconds per execution.

\subsubsection{Baselines and ZhuLong Variants}

To isolate the contribution of each component, we establish two baselines and three ZhuLong variants, as shown in Table~\ref{tab:configurations}. ZhuLong variants default to vector retrieval using a name+description index.

\begin{table}[htbp]
\centering
\caption{Configurations and their components}
\label{tab:configurations}
\begin{tabular}{lccc}
\toprule
\textbf{Configuration} & \textbf{Retrieval} & \textbf{Sandbox} & \textbf{Self-Expl.} \\
\midrule
Baseline 1 (Pure LLM) & \ding{55} & \ding{55} & \ding{55} \\
Baseline 2 (RAG) & \ding{51} & \ding{55} & \ding{55} \\
ZhuLong w/o Self-Expl. & \ding{51} & \ding{51} & \ding{55} \\
ZhuLong w/o Sandbox & \ding{51} & \ding{55} & \ding{51} \\
ZhuLong w/o Retrieval & \ding{55} & \ding{51} & \ding{51} \\
\textbf{ZhuLong} (Full) & \ding{51} & \ding{51} & \ding{51} \\
\bottomrule
\end{tabular}
\end{table}

\subsubsection{LLM Backbones}

All main experiments use DeepSeek-V4-Flash as the primary backbone. To understand the impact of different LLMs, we additionally evaluate DeepSeek-V3.2, GLM-5.1, DeepSeek-V4-Pro, Kimi-K2.6, and Doubao-Seed-2.0-Pro.

\subsection{Main Results and Ablation}
\label{sec:exp-results}

We further ablate the number of re-planning rounds $n \in \{0, 1, 2\}$.

Table~\ref{tab:main-ablation} presents the progressive build-up (top), the
iteration budget ablation (middle), and the component ablation (bottom).
The top and bottom blocks use the same re-planning budget of 2 rounds but
present the results from complementary perspectives: the top block shows
the sequential gains from incrementally adding components, with each row
compared against the previous one, whereas the bottom block shows the
performance degradation caused by removing individual components from the
full system. The middle block isolates the effect of the re-planning budget
by comparing 0, 1, and 2 rounds within the full system.

\begin{table}[htbp]
\centering
\caption{Ablation study results}
\label{tab:main-ablation}
\begin{tabular}{lcc}
\toprule
\textbf{Configuration} & \textbf{Pass@1 (\%)} & \textbf{$\Delta$} \\
\midrule
\multicolumn{3}{l}{\textit{Progressive build-up (vs. previous row)}} \\
Baseline 1: Pure LLM & 23.6 & — \\
Baseline 2: RAG & 32.3 & +8.7 pp \\
ZhuLong w/o Self-Expl. & 75.3 & +43.0 pp \\
\textbf{ZhuLong} (Full) & \textbf{78.5} & \textbf{+3.2 pp} \\
\midrule
\multicolumn{3}{l}{\textit{Iteration budget ablation (vs. ZhuLong Full)}} \\
0 re-planning rounds (single attempt) & 62.7 & -15.8 pp \\
1 re-planning round & 76.0 & -2.5 pp \\
\midrule
\multicolumn{3}{l}{\textit{Component ablation (vs. ZhuLong Full)}} \\
ZhuLong w/o Self-Expl. & 75.3 & -3.2 pp \\
ZhuLong w/o Sandbox & 37.3 & -41.2 pp \\
ZhuLong w/o Retrieval & 34.2 & -44.3 pp \\
\bottomrule
\end{tabular}
\end{table}

\subsubsection{Effect of Iteration Budget}
\label{sec:iteration-budget-effect}

The middle block of Table~\ref{tab:main-ablation} isolates the iteration budget.
With 0 re-planning rounds, ZhuLong achieves 62.7\% Pass@1, already far above 
the 32.3\% RAG baseline.
One re-planning round raises accuracy to 76.0\% (+13.3 pp, 84.2\% of the total 
achievable gain within two iterations), while the second round adds only 
2.5 pp (final 78.5\%).
This diminishing return indicates that most recoverable errors are corrected 
within one round; we retain two rounds as the default to capture the remaining 
marginal improvement without excessive overhead.

\subsubsection{Effect of Sandbox Execution (RQ1)}

Sandbox execution is the dominant performance driver. Adding it to the RAG baseline raises Pass@1 from 32.3\% to 75.3\%---a gain of \textbf{43.0 percentage points} that accounts for \textbf{78.3\%} of the total improvement from Pure LLM (23.6\%) to ZhuLong (78.5\%). Conversely, removing sandbox execution from ZhuLong drops Pass@1 to 37.3\%, a decline of \textbf{41.2 percentage points}. These results confirm that execution feedback is indispensable: without it, the agent cannot verify correctness, observe API behavior, or iteratively refine its output.

We attribute this to two mechanisms. First, execution provides concrete error messages that guide corrections, transforming one-shot generation into iterative search. Second, the agent can experiment with parameter combinations in the sandbox, compensating for gaps in documentation \cite{kumar2026agentforge, jiang2024seed}.

\subsubsection{Effect of Self-Exploration (RQ2)}

Removing the self-exploration mechanism (while keeping sandbox execution and retrieval) reduces Pass@1 from ZhuLong's 78.5\% to 75.3\%, a drop of 3.2 percentage points. This confirms that the self-exploration mechanism contributes a measurable accuracy gain beyond sandbox execution.

Beyond accuracy, the self-exploration mechanism also improves efficiency. Table~\ref{tab:selfdoc-efficiency} compares tool usage with and without self-exploration. With self-exploration enabled (i.e., ZhuLong), the agent produces more traces (243 vs. 220) because it persists longer with better documentation, yet total tool calls drop by 13.9\% and average calls per trace decrease by 22.1\%.

\begin{table}[htbp]
\centering
\caption{Overall tool usage: with vs. without self-exploration}
\label{tab:selfdoc-efficiency}
\begin{tabular}{lcc}
\toprule
\textbf{Metric} & \textbf{w/o Self-Expl.} & \textbf{ZhuLong (Full)} \\
\midrule
Traces & 220 & 243 (+10.5\%) \\
\#Calls & 6,791 & 5,845 (-13.9\%) \\
Avg/Trace & 30.9 & 24.1 (-22.1\%) \\
\midrule
Change & — & \textbf{-946 (-13.9\%)} \\
\bottomrule
\end{tabular}
\end{table}

The reduction is most pronounced in \texttt{run\_code} (27.0\%, see Table~\ref{tab:tool-breakdown}), indicating that enriched documentation reduces trial-and-error executions. Calls to \texttt{search\_apis} decrease by 12.2\%, while \texttt{get\_api\_details} increase by 9.0\%---a behavioral shift toward more thorough reading before coding.

\begin{table}[htbp]
\centering
\caption{Tool-wise call count breakdown}
\label{tab:tool-breakdown}
\begin{tabular}{lcc}
\toprule
\textbf{Tool} & \textbf{w/o Self-Expl.} & \textbf{ZhuLong (Full)} \\
\midrule
\texttt{search\_apis} & 3,203 & 2,812 (-12.2\%) \\
\texttt{get\_api\_details} & 1,151 & 1,255 (+9.0\%) \\
\texttt{run\_code} & 2,437 & 1,778 (-27.0\%) \\
\bottomrule
\end{tabular}
\end{table}

In the absence of sandbox execution, the self-exploration mechanism alone improves Pass@1 from 32.3\% (RAG) to 37.3\% (ZhuLong w/o Sandbox), a gain of 5.0 percentage points. This confirms that enhanced documentation improves generation accuracy even without iterative refinement. The larger gain in the no-sandbox setting (5.0 pp vs. 3.2 pp in ZhuLong) suggests that sandbox execution partially substitutes for pre-explored documentation.

\subsubsection{Retrieval Strategy (RQ3) and Index Construction (RQ4)}

Vector-based semantic retrieval outperforms grep-based keyword search by 12.7 percentage points (78.5\% vs. 65.8\%), confirming that natural language queries for EDA tasks rarely match exact API names. For index construction, the name+description index achieves the best result (78.5\%), outperforming description-only (76.0\%) and name-only (75.3\%), indicating that names enable exact matching while descriptions support semantic generalization. Prompt language (English vs. Chinese) shows no material difference, both achieving 78.5\% Pass@1.

\subsubsection{Impact of LLM Backbone}

Table~\ref{tab:llm-comparison} presents the Pass@1 results for six LLM backbones integrated into the full ZhuLong system. We adopt DeepSeek‑V4‑Flash as the primary backbone across all experiments in this paper, based on a practical trade‑off between model capability and inference cost: its performance is sufficiently strong to support the key findings from both our main experiments and ablation studies, while its cost remains manageable for running the complete set of configurations reported in this work. Kimi‑K2.6 achieves the highest accuracy (83.5\%), outperforming DeepSeek‑V4‑Flash (78.5\%) by 5.0 percentage points. DeepSeek‑V4‑Pro also surpasses this baseline, reaching 81.0\% (+2.5 pp), while GLM‑5.1 ties with DeepSeek‑V4‑Flash at 78.5\%. In contrast, DeepSeek‑V3.2 and Doubao‑Seed‑2.0‑Pro lag notably behind, yielding 67.1\% and 55.7\%, respectively. The spread of 27.8 percentage points between the best and worst performers indicates that the underlying model remains a decisive factor, even with retrieval, sandbox, and self‑exploration in place. Notably, the ranking on this benchmark does not strictly follow the general coding proficiency reported in public leaderboards. This discrepancy suggests that EDA scripting—with its long‑tail APIs and state‑sensitive logic—imposes a distinct set of demands that generic code benchmarks do not adequately capture.

\begin{table}[htbp]
\centering
\caption{Impact of LLM backbone (ZhuLong)}
\label{tab:llm-comparison}
\begin{tabular}{lcc}
\toprule
\textbf{LLM Backbone} & \textbf{Pass@1 (\%)} & \textbf{$\Delta$} \\
\midrule
DeepSeek-V4-Flash & 78.5 & — \\
GLM-5.1 & 78.5 & — \\
DeepSeek-V3.2 & 67.1 & -11.4 pp \\
DeepSeek-V4-Pro & 81.0 & +2.5 pp \\
Kimi-K2.6 & 83.5 & +5.0 pp \\
Doubao-Seed-2.0-Pro & 55.7 & -22.8 pp \\
\bottomrule
\end{tabular}
\end{table}

\subsection{Error Analysis}
\label{sec:error-analysis}

ZhuLong produces correct code for 124 of 158 tasks (78.5\% Pass@1). The 
remaining 34 failures comprise 13 generation-stage failures and 21 
execution-stage failures. The generation failures fall into two categories: 
the majority are timeouts on lengthy, multi-step tasks, reflecting 
computational budget constraints rather than capability limits, as traces 
show ongoing reasoning at termination; in a few cases, the agent exhausted 
its message budget cycling through \texttt{search\_apis} calls, unable to 
locate the correct API from natural-language queries. 
We manually inspected 
all 21 execution-stage failure traces and identify two root cause categories 
(Table~\ref{tab:error-categories}).

\begin{table}[t]
\centering
\caption{Root cause categories for execution-stage failures}
\label{tab:error-categories}
\begin{tabular}{lcc}
\toprule
Category & Count & Example Tasks \\
\midrule
API misuse & 12 & 068, 074, 092, 098, 063, \\ 
& & 117, 133, 134, 121, 115, \\
& & 114, 118 \\
\midrule
API composition and  & 9 & 051, 058, 099, 111, 146, \\
workflow errors & & 149, 150, 158, 052 \\
\bottomrule
\end{tabular}
\end{table}

\subsubsection{API misuse.}

The agent selects the correct API family but violates undocumented API contracts—such as hidden type expectations, internal state dependencies, or context preconditions—that the C++-generated bindings enforce but the documentation does not reveal.

\textit{Case study: Task 098 (delete\_design\_figures).} The agent located \texttt{aeSelectFigs} and \texttt{aeDeleteObj} by searching for ``delete figure objects,'' then passed a plain Python list to \texttt{aeSelectFigs}. The call succeeded, but the subsequent \texttt{aeDeleteObj} failed because the internal C++ selection state—which the API relies on—was never properly populated by the plain Python list. Across five subsequent \texttt{run\_code} calls within the same trace,
the agent modified only its self-generated test harness rather than
correcting the underlying type mismatch, as the runtime diagnostic
(``Error: Failed to delete the selected figures.'') provided no
actionable signal.

Other instances share similar type-invisibility patterns: Task~074 passed 
a Python list where \texttt{CStringList} was expected; Task~063 called a 
function that returned an internal type name such as \texttt{"maskLayout"}, 
while the test harness used a different getter that produced 
\texttt{"cellFileType"}; and Task~114 selected \texttt{aeCreateRect}, which 
requires an active AE GUI context, instead of \texttt{dbCrtRect(cv, ...)} 
that operates directly on a design handle. The remaining cases in this 
category follow analogous API selection or type mismatches.

\subsubsection{API composition and workflow errors.}

The agent selects individually valid APIs, and the generated code executes without runtime errors, but the composition logic---parameter propagation, post-processing, or control flow---fails the test oracle. These failures are harder to detect because each call appears correct in isolation.

\textit{Case study: Task 052 (check\_view\_list).} The specification required invoking \texttt{EmyNtlUtil.checkViewList} for netlist view validation. Instead, the agent implemented a custom validator using pure Python logic. The code printed ``All assertions passed'' in its own embedded tests, but the harness applied standard validation on a different configuration, where the heuristic returned an error message rather than \texttt{"VALID"}. The agent replaced the mandated library call with an approximation, violating the test assertion.

Other instances include: Task~058, where the agent used substring matching (\texttt{"high" in queue\_name}) instead of the equality check (\texttt{queue\_name == "priority"}) required by the specification, producing wrong \texttt{misc} parameter values; Task~051, where \texttt{str(result).strip()} caused an assertion-level type mismatch against the test's \texttt{QString} expectation; and Task~099, where the agent used \texttt{emyPointArrayShrinkGrow} to 
expand a polygon, but the correct API was \texttt{dbExpandPoints} (bounding-box 
expansion). The remaining cases exhibit similar errors, where the agent either 
composed valid APIs incorrectly or substituted a mandated library call with 
a custom approximation.


\subsubsection{Implications.}

The analysis points to two recurring causes of execution-stage failures.
First, undocumented API contracts and binding-level constraints are
unavailable during code generation, forcing the agent to guess argument
formats, while subsequent sandbox feedback often fails to expose the
underlying constraints. Second, missing usage-level knowledge---such as
correct API combinations, parameter conventions, and common pitfalls---
forces the agent to rediscover solutions through trial and error. These
findings motivate type-aware API documentation and case-level knowledge
distillation, where resolved failures are stored as reusable experiences,
forming the basis of agent self-evolution.


\subsection{Interactive Scenario Evaluation}
\label{sec:exp-interactive}

Beyond standalone execution tasks, real-world EDA scripting often involves direct manipulation of unsaved layouts and schematics. To assess ZhuLong in such GUI-bridged scenarios, we collected 20 tasks requiring operation on volatile in-memory states, window focus handling, and GUI-level interactions. All tasks were executed via the interactive mode of \texttt{run\_code} with a single attempt (no retry), as the iterative recovery loop is not supported in the bridge-based environment. These evaluations were conducted with self-exploration disabled; results thus reflect ZhuLong w/o Self-Exploration in interactive settings.

Table~\ref{tab:interactive-results} shows the results across the same 20 tasks for both PyAether and SKILL. The overall Pass@1 is \textbf{60.0\%} for PyAether and \textbf{50.0\%} for SKILL—notably lower than the 78.5\% achieved on the main benchmark. This gap is attributed to two factors: (1) no iterative recovery loop, limiting each task to a single attempt; and (2) self-exploration disabled. Performance varies by task category: PyAether excels in focus-dependent and modification tasks (75.0\% or higher), while SKILL shows an advantage in design content manipulation and hierarchical inspection. Focus-dependent operations are the most challenging for SKILL (25.0\%), while design content manipulation is the most difficult for PyAether (37.5\%). The 10~pp overall gap reflects Python's more favorable syntax for LLM-based synthesis compared to SKILL's Lisp-based prefix notation.

\begin{table}[htbp]
\centering
\caption{Interactive scenario results: same 20 tasks for PyAether and SKILL}
\label{tab:interactive-results}
\begin{tabular}{lccc}
\toprule
\textbf{Category} & \textbf{Tasks} & \textbf{PyAether} & \textbf{SKILL} \\
\midrule
Design content manipulation & 8 & 37.5\% & 50.0\% \\
Focus-dependent operations & 4 & 75.0\% & 25.0\% \\
Hierarchical inspection & 4 & 50.0\% & 75.0\% \\
Design modification & 2 & 100.0\% & 50.0\% \\
GUI interaction & 2 & 100.0\% & 50.0\% \\
\midrule
Overall & 20 & 60.0\% & 50.0\% \\
\bottomrule
\end{tabular}
\end{table}
\section{Discussion}
\label{sec:discussion}

The error analysis in Section~\ref{sec:error-analysis} shows that the 21 execution-stage failures mainly arise from API misuse and incorrect API composition. These failures are largely caused by undocumented API contracts, missing usage-level knowledge, and insufficient validation feedback, motivating type-aware retrieval and more informative execution feedback.

Beyond these specific improvements, our results carry broader implications. The dominance of sandbox execution (+43 pp) over static knowledge suggests that for domains with long-tailed, under-docu\-mented APIs, execution grounding is not merely beneficial but essential. This paradigm may generalize to other domain-specific scripting tasks with similar characteristics.

Several limitations also warrant acknowledgment. First, performance varies significantly across base LLMs (55.7\%–83.5\%), indicating that the underlying model remains a decisive factor. Second, offline self-exploration incurs upfront cost (one-time per API), amortized over subsequent queries. Third, the interactive bridge is single-threaded, limiting concurrent deployment. Fourth, while realistic, the benchmark is limited to 158 PyAether tasks; expansion to SKILL and other EDA platforms remains future work.
\section{Conclusion and Future Work}
\label{sec:conclusion}

This paper presented \textbf{ZhuLong}, an execution-grounded LLM coding agent for PyAether and SKILL that closes the loop from generation to execution to iterative refinement via sandbox execution feedback. Evaluated on \textbf{EDA-Eval-PyAether}—the first benchmark for PyAether scripting, comprising 158 real-world tasks—the complete system achieves 78.5\% Pass@1, substantially outperforming static RAG (32.3\%). Ablation studies identify execution feedback as the dominant performance driver (+43 pp), with offline API self-exploration providing complementary gains in both accuracy (+3.2 pp) and efficiency (22.1\% fewer tool calls per trace). These results demonstrate that, for domain-specific scripting with long-tail, under-documented APIs, execution grounding is not merely beneficial but essential.

Beyond the specific results, this work establishes two foundations for future research: (1) a \textbf{methodological paradigm}—execution-grounded agents with offline API exploration—that can generalize to other domains with similar characteristics (long-tail APIs, incomplete documentation, available execution sandboxes); and (2) an \textbf{evaluation infrastructure}—EDA-Eval-PyAether—that enables systematic comparison and progress tracking for LLM-based EDA scripting. We will release both code and benchmark upon acceptance.


Future work proceeds along three directions. First, extending the interactive benchmark to systematically cover diverse GUI operations and volatile environment states, and enabling self-exploration in interactive settings. Second, evolving the agent through self-improvement—learning from both successful and failed execution traces to refine its retrieval, code generation, and exploration strategies. Third, adapting the architecture to additional EDA platforms, including Synopsys tools with Tcl scripting, to broaden the scope of execution-grounded EDA agents.


\bibliographystyle{ACM-Reference-Format}
\bibliography{reference}

\end{document}